\documentclass[sigconf]{acmart}
\usepackage{makecell}

\renewcommand\footnotetextcopyrightpermission[1]{}

\def\BibTeX{{\rm B\kern-.05em{\sc i\kern-.025em b}\kern-.08emT\kern-.1667em\lower.7ex\hbox{E}\kern-.125emX}}
    
\copyrightyear{2020} 
\acmYear{2020} 
\setcopyright{acmcopyright}
\acmConference[CoDS COMAD 2020]{7th ACM IKDD CoDS and 25th COMAD}{January 5--7, 2020}{Hyderabad, India}
\acmBooktitle{7th ACM IKDD CoDS and 25th COMAD (CoDS COMAD 2020), January 5--7, 2020, Hyderabad, India}
\acmPrice{15.00}
\acmDOI{10.1145/3371158.3371196}
\acmISBN{978-1-4503-7738-6/20/01}

\begin{document}
\sloppy

\title{PlantDoc: A Dataset for Visual Plant Disease Detection}
\author{Davinder Singh*, Naman Jain*, Pranjali Jain*, Pratik Kayal*}\thanks{*Equal Contribution}
\author{Sudhakar Kumawat, Nipun Batra}
\affiliation{Indian Institute of Technology Gandhinagar, Gujarat, India 382 355}
\email{{davinder.singh, naman.j, pranjali.jain, pratik.kayal, sudhakar.kumawat, nipun.batra}@iitgn.ac.in}

\renewcommand{\shortauthors}{D. Singh*, N. Jain*, P. Jain*, P. Kayal*, S. Kumawat, N. Batra}

\begin{abstract}
India loses 35\% of the annual crop yield due to plant diseases. Early detection of plant diseases remains difficult due to the lack of lab infrastructure and expertise. In this paper, we explore the possibility of computer vision approaches for scalable and early plant disease detection. The lack of availability of sufficiently large-scale non-lab data set remains a major challenge for enabling vision based plant disease detection. Against this background, we present PlantDoc: a dataset for visual plant disease detection.
Our dataset contains 2,598 data points in total across 13 plant species and up to 17 classes of diseases, involving approximately 300 human hours of effort in annotating internet scraped images. To show the efficacy of our dataset, we learn 3 models for the task of plant disease classification. Our results show that modelling using our dataset can increase the classification accuracy by up to 31\%. We believe that our dataset can help reduce the entry barrier of computer vision techniques in plant disease detection.

\end{abstract}

\maketitle

\section{Introduction}
% ~\citeN{oerke2012crop} 
% For many people on the planet, food is a given. However, for the staggering 821 million people who are hungry, food is not a guarantee. 
% Civilizations flourished through the steady supply of food possible due to enormous human effort. With consistent growth and innovation in the healthy food supply,

% first paragraph is the motivation: why work on plant diseases? (first line of abstract)
% second paragraph is related work: why existing techniques don't work (2nd line of abstract): they require lab, domain expertise, costly and not scalable
% 3rd para is the main idea: computer vision has made an impact in other fields...then also mention about the role of datasets like Imagenet. Say no such dataset exists for plant diseases. Existing ones are controlled or very small.
% 4th para is details of your datasets. Put the pertinent numbers.
% 5th para is about experiments you did to show the efficacy of your dataset

\begin{figure*}[t]
\begin{tabular}{c|ccccccc}
 & \makecell{Apple \\ Black Rot} & \makecell{Bell Pepper \\Bacterial} & \makecell{Blueberry \\ Healthy} & \makecell{Cherry \\Powdery Mildew} & \makecell{Corn \\ Gray Spots} & \makecell{Grape \\ Black Rot} & \makecell{Potato \\ Early Blight} \\
\centering\textbf{PVD} & \includegraphics[height=1.4cm]{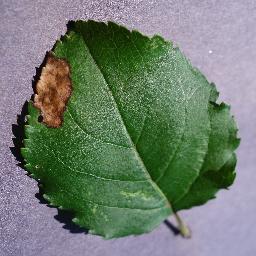} & \includegraphics[height=1.4cm]{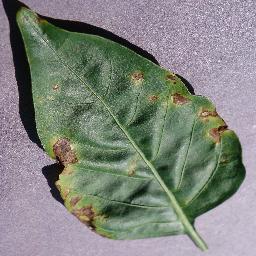} & \includegraphics[height=1.4cm]{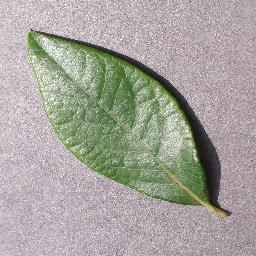} & \includegraphics[height=1.4cm]{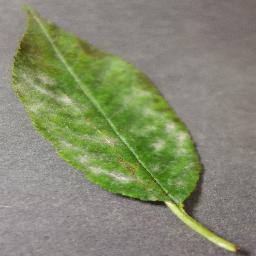} & \includegraphics[height=1.4cm]{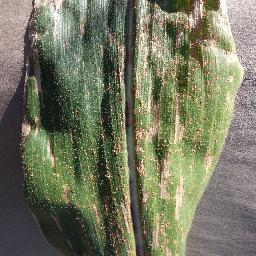} & \includegraphics[height=1.4cm]{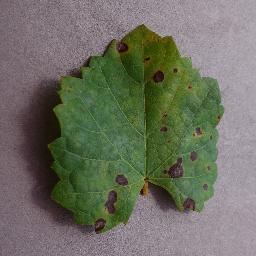} &  \includegraphics[height=1.4cm]{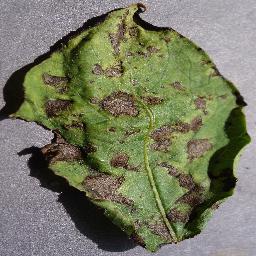} \\
\textbf{PlantDoc}& \includegraphics[height=1.4cm]{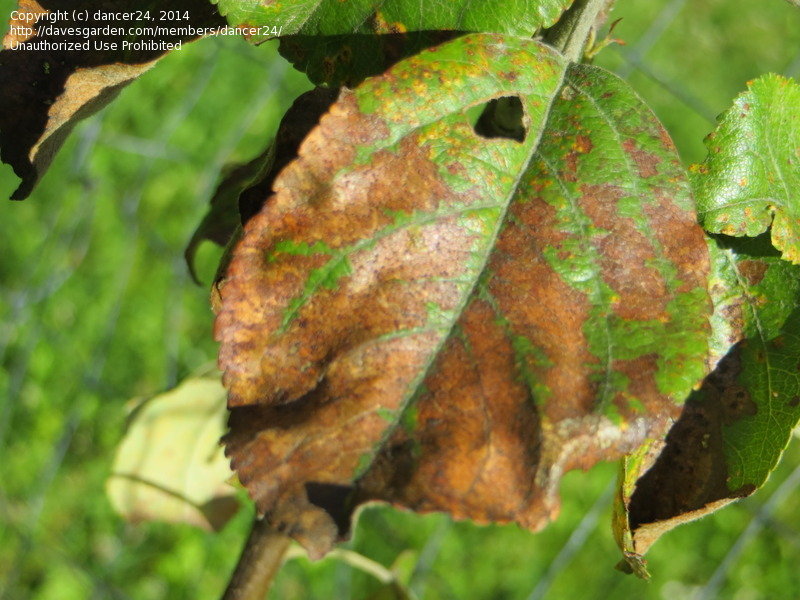} & \includegraphics[height=1.4cm]{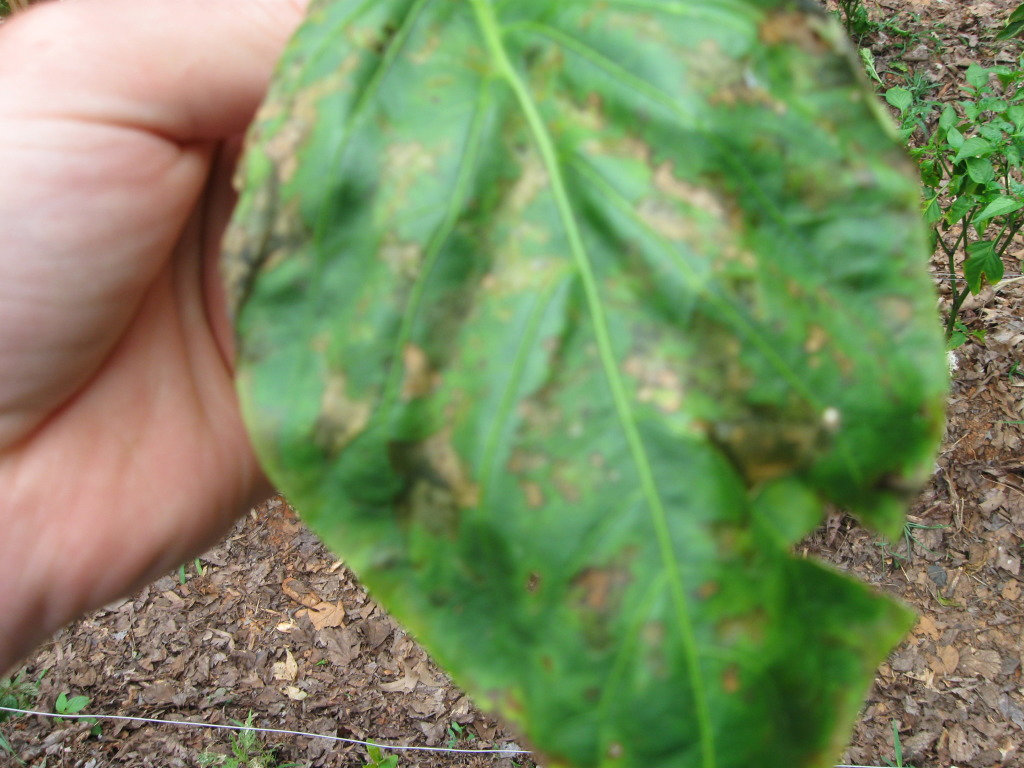} & \includegraphics[height=1.4cm]{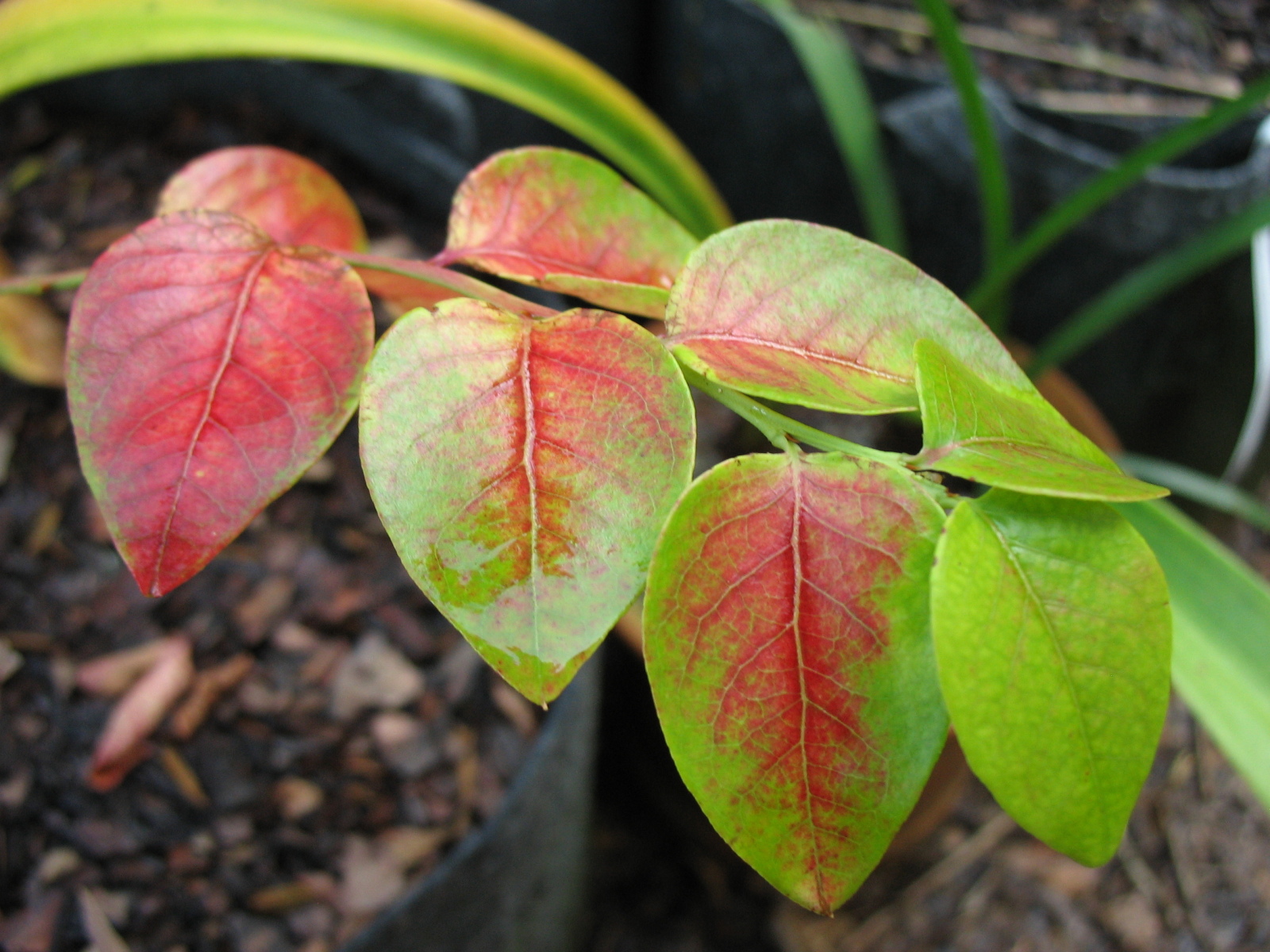} & \includegraphics[height=1.4cm]{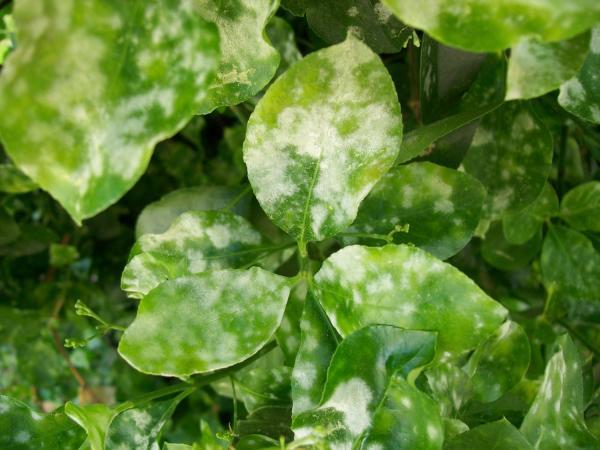} & \includegraphics[height=1.4cm]{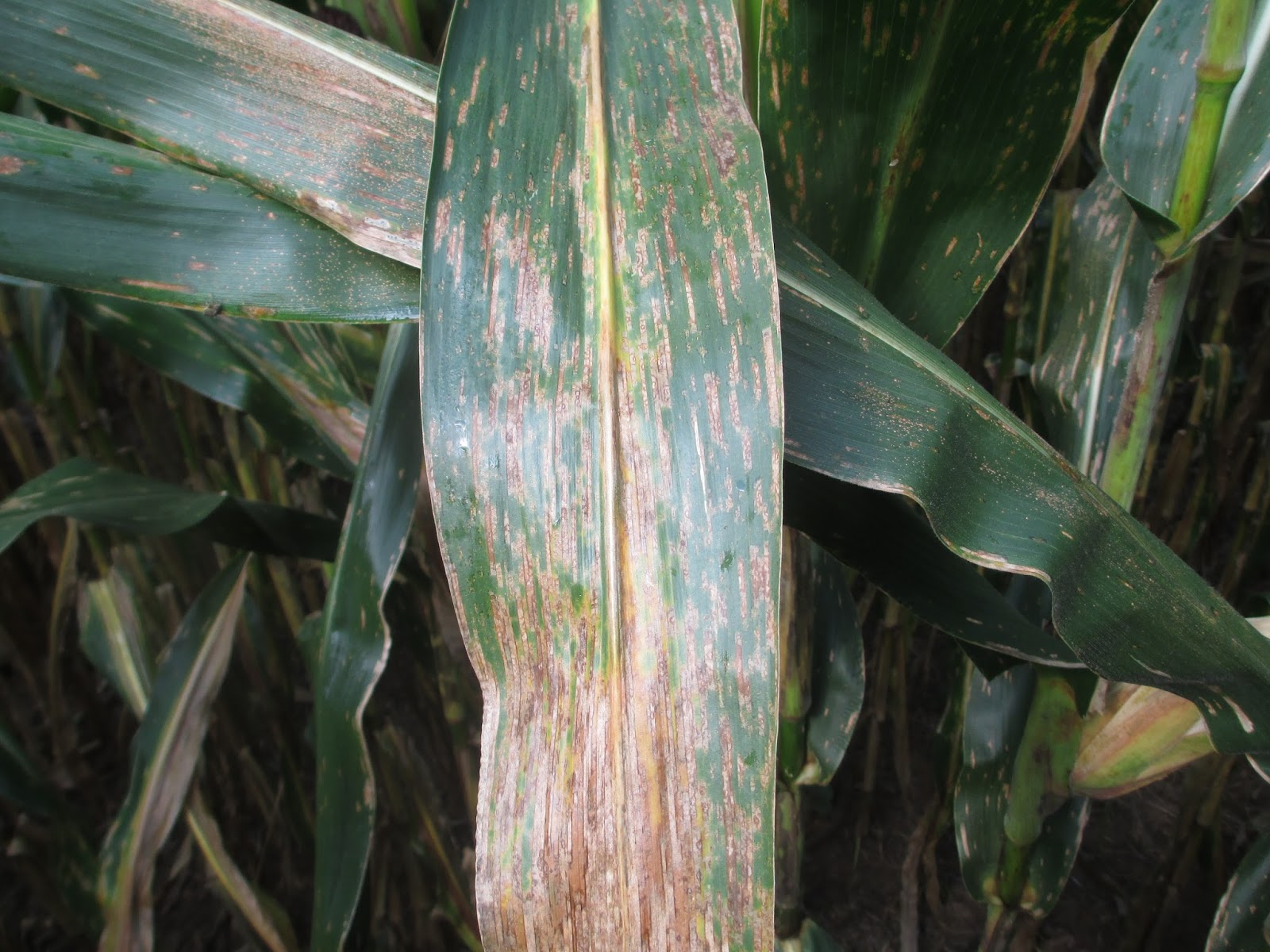} & \includegraphics[height=1.4cm]{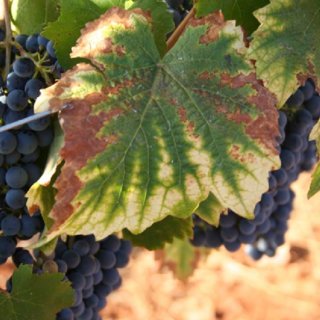} &  \includegraphics[height=1.4cm]{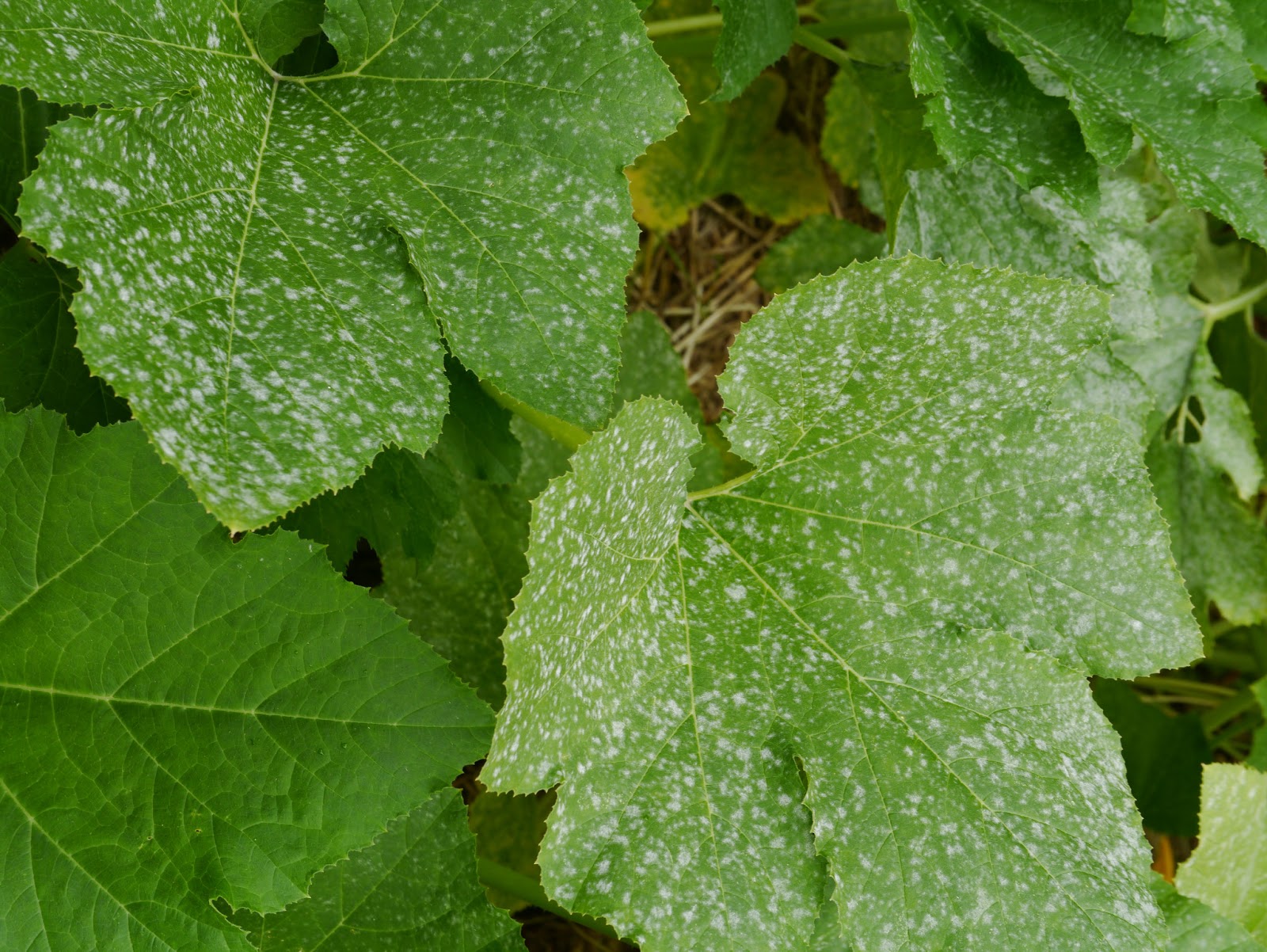}
\end{tabular}
\caption{Samples from various classes in the PlantDoc Dataset show the gap between lab-controlled and real-life images}\label{fig:dataset}
\end{figure*}

Annually the Earth's population increases by about 1.6\%, and so does the demand for plant products of every kind ~\cite{oerke2012crop}. The protection of crops against plant diseases has a vital role to play in meeting the growing demand for food quality and quantity ~\cite{strange2005plant}. In terms of economic value, plant diseases alone cost the global economy around US\$220 billion annually \cite{agrios2005plant}. According to the Indian Council of Agricultural Research, more than 35\% of crop production is lost every year due to Pests and Disease ~\cite{ICARNewsletter}. Food security is threatened by an alarming increase in the number of outbreaks of pests and plant diseases. These diseases jeopardize food security and have broad economic, social, and environmental impacts \cite{dusseldorfsave}. 

Timely disease detection in plants remains a challenging task for farmers. They do not have many options other than consulting fellow farmers or the Kisan helpline \cite{kisan}. Expertise in plant diseases is necessary for an individual to be able to identify the diseased leaves. Furthermore, in most cases it is necessary to have a lab infrastructure to identify a diseased leaf.  

In this work, we explore the possibility of using computer vision for scalable and cost-effective plant disease detection. Computer vision has made tremendous advances in the past few years through various advances in deep convolutional neural networks. While training large neural networks can be very time consuming, the trained models can classify images very quickly, which makes them also suitable for consumer applications on smartphones. Image processing for detecting plant diseases opens up new avenues to combine the knowledge of deep learning approaches with real-world problems in agriculture, and hence, facilitates advancements in agricultural knowledge, the yield of crops, and disease control. 

Majority of existing vision-based solutions require high-resolution images with a plain background. In contrast, as the majority of Indian farmers use low-end mobile devices with natural background and lighting conditions, we focus on images in natural environmental conditions with non-trivial background noise and provide the best possible query resolution for crops and plants. Against this background, we highlight our two main contributions: i) development of PlantDoc: a dataset of 2,598 images across 13 plant species and 27 classes(17-10, disease-healthy) ii) benchmarking the curated data set and showing its utility in disease detection in non-controlled environments. To the best of our knowledge, this is the first such dataset containing data from non-controlled settings. 

We evaluated our dataset using various classification and object detection architectures mentioned in Section \ref{sec:experiment} to establish the requirement of a dataset in non-controlled settings. The results suggested that lab-controlled dataset cannot be used to classify or detect images in real-scenario. We found that fine-tuning the models on PlantDoc reduces the classification error by up to 31\%. Thus, our dataset can potentially be used to build an application which detects and classifies 27 plant disease/healthy classes efficiently.

\section{Related Work} \label{sec:prior}
Our related work can be broadly categorized into: i) techniques for plant disease detection; and ii) datasets advancing research in plant disease detection. 

\subsection{Techniques for plant disease detection}
Prior work by Sankaran et al.~\cite{sankaran2010review} proposed using reliable sensors for monitoring health and diseases in plants under field conditions. However, plant disease detection using sensors has the potential to benefit only a few farmers because of the substantial hardware cost and lack of expertise to operate such sensors. In contrast, prior work by Patil et al.~\cite{patil2011leaf} extracted shape features for disease detection in sugarcane leaves obtaining a final average accuracy of 98.60\%. In a similar work, Patil et al.~\cite{patilsanjay} used texture features, namely inertia, homogeneity, and correlation obtained by calculating the gray level co-occurrence matrix on the image and color extraction for disease detection on maize leaves. Recent work~\cite{grinblat2016deep} has looked into neural networks for the identification of three different legume species based on the morphological patterns of leaves veins. Likewise, feature extraction and Neural Network Ensemble (NNE) have been used for recognizing tea leaf diseases with a final testing accuracy of 91\%~\cite{zhou2002neural}. A host of other recent works have looked at convolutional neural network variants for disease detection using plant leaf images~\cite{sladojevic2016deep,fuentes2017robust}. These works are limited to a particular crop, which is a significant limitation. Also, the datasets used in the works have not been made public, thereby, impacting reproducibility.

\subsection{Datasets for plant disease detection}

The PlantVillage dataset(PVD)~\cite{mohanty2016} is the only public dataset for plant disease detection to the best of our knowledge. The data set curators created an automated system using GoogleNet \cite{szegedy2015going} and AlexNet \cite{krizhevsky2012imagenet} for disease detection, achieving an accuracy of 99.35\%. However, the images in PlantVillage dataset are taken in laboratory setups and not in the real conditions of cultivation fields, due to which their efficacy in real world is likely to be poor. In contrast, we curate real-life images of healthy and diseased plants to create a publicly available dataset.

\section{The PlantDoc Dataset}
 \begin{figure}[h]
\centering
\includegraphics[width=\linewidth]{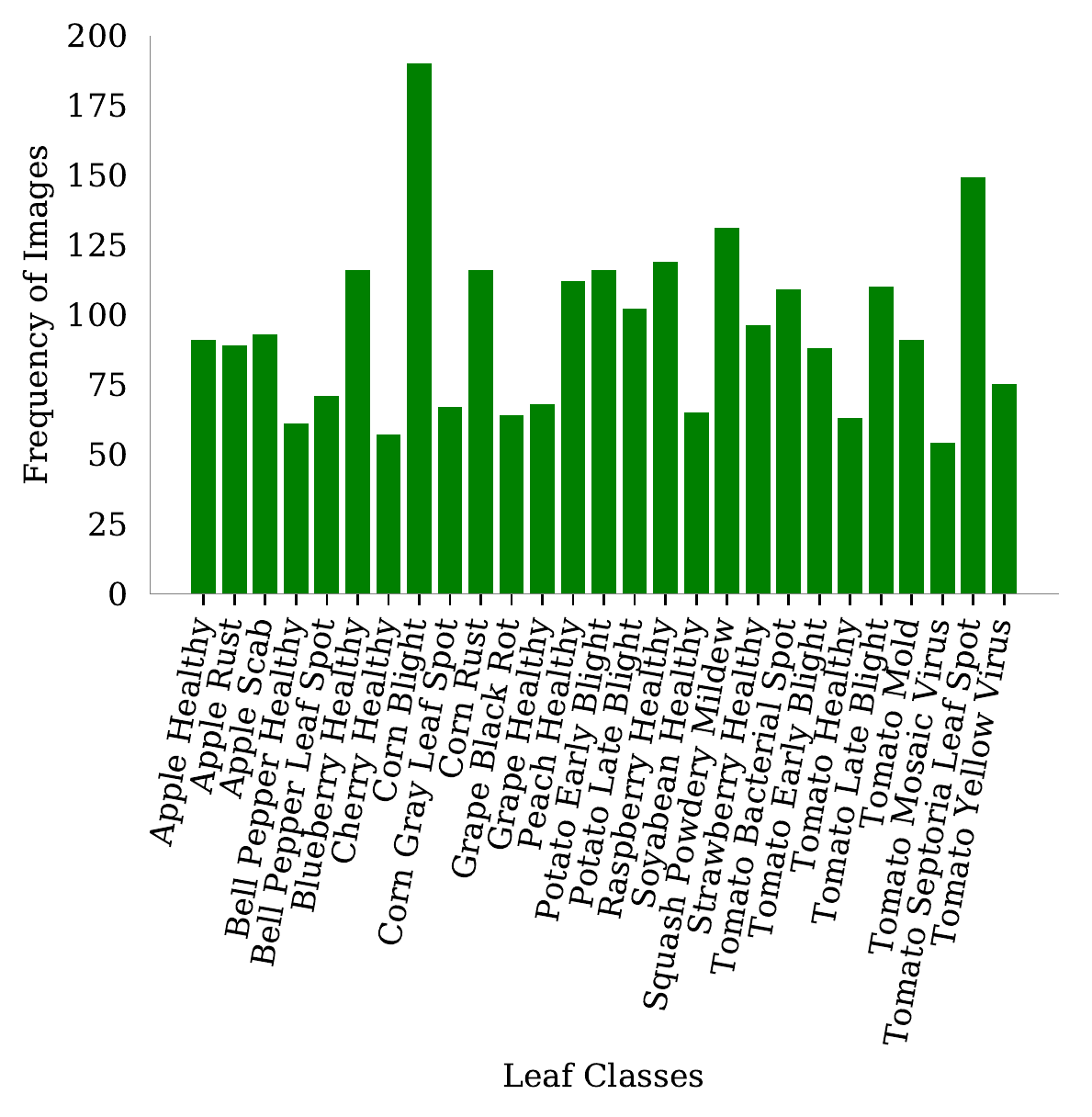}
\caption{Statistics of PlantDoc Dataset}\label{fig:isd}
\end{figure}

\label{sec:method}
The PlantVillage dataset contains images taken under controlled settings. This dataset limits the effectiveness of detecting diseases because, in reality, plant images may contain multiple leaves with different types of background conditions with varying lighting conditions (shown in Figure \ref{fig:dataset}). Against this background, we now describe our curated dataset and discuss the techniques used for curation.
%  \begin{figure}[t]
% \centering
% \resizebox{\hsize}{!}{\begin{tabular}{c}
%      \includegraphics{fig/bar_2.pdf}
% \end{tabular}}
% \caption{Statistics of PlantDoc Dataset}\label{fig:isd}
% \end{figure}

\subsection{Data Collection}
To account for the intricacies of the real world, we require models trained on real-life images. This fact motivated us to create a dataset by downloading images from Google Images and Ecosia~\cite{ecosia} for accurate plant disease detection in the farm setting. We downloaded images from the internet since collecting large-scale plant disease data through fieldwork requires enormous effort. We collected about 20,900 images by using scientific and common names of 38 classes mentioned in the dataset by \citet{mohanty2016}. 

\begin{figure}[t]
\centering
\resizebox{0.8\hsize}{!}{\begin{tabular}{c}
    \includegraphics{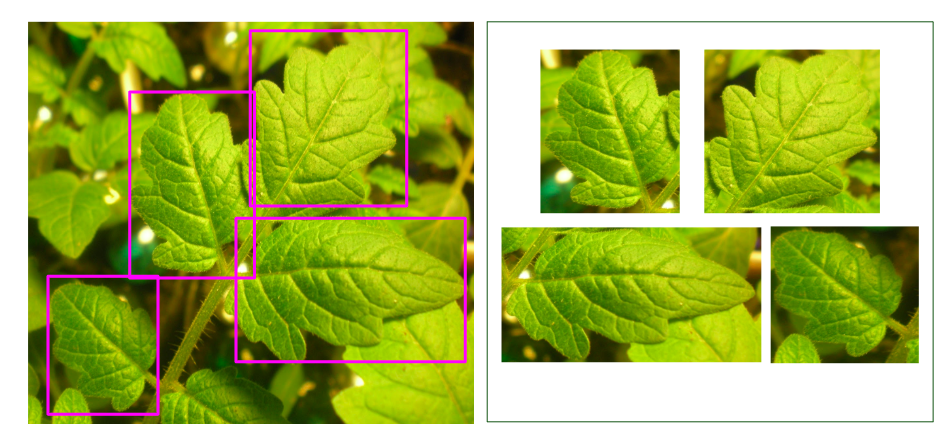}
\end{tabular}}
\caption{An Image with bounding boxes and its cropped leaves}\label{fig:boundbox}
\end{figure}

Four users filtered the images by selecting images based on their metadata on the website and guidelines mentioned on APSNet \cite{apsnet}. APS compiled a list of peer-reviewed literature corresponding to each plant disease. We referred APS' prior literature and accordingly classified images. Some of the most important factors for classification were the color, area and density of the diseased part and shape of the species.
We removed inappropriate (such as non-leaf plant, lab controlled and out-of-scope images) and duplicate images across classes downloaded due to web search. Every image was checked by two individuals according to the guidelines to reduce labeling errors. Finally, to have sufficient training samples, we removed the classes with less than 50 images. Figure \ref{fig:isd} shows the statistics of the final dataset having a total of 27 classes spanning over 13 species with 2,598 images.
To build an application for the object detection task, we need exact bounding regions containing the leaf in the entire image. Hence, we used the \texttt{LabelImg} tool \cite{tzutalin2015labelimg} to make the bounding boxes around the leaves (Figure \ref{fig:boundbox}) in all the images. In real scenarios, the image may have multiple leaves or a combination of diseased and healthy leaves. We labeled all the leaves in the image explicitly with their particular classes. While labeling the boxes, we made sure that the entire leaf should be present inside the box and the area of the bounding box should not be smaller than 1/8th (approximately) of the image size. After labeling, the information about all the coordinates of boxes in an image and their respective class label were stored separately in an XML file corresponding to each image.

\noindent \textbf{Cropped-PlantDoc Dataset}:
To show the differences between our dataset and PlantVillage, we built another dataset called the \textit{Cropped-PlantDoc (C-PD)}  by cropping the images using bounding box information. Similar to PlantVillage, cropped images contains only the leaf but these images are of low-quality, have small-size and varying backgrounds. The total number of leaf images after cropping 2,598 images turns out to be 9,216 i.e. 9,216 bounding boxes.  

\section{Benchmarking PlantDoc Dataset}
\label{sec:experiment}
We now discuss two benchmark set of experiments on our dataset: i) plant image classification; and ii) detecting leaf within an image. 

\begin{table}[!thb]
\begin{center}
\resizebox{\hsize}{!}{
\begin{tabular}{ccc|cc}
% \hline
\bf PreTrained Weights & \bf Training Set & \bf Test Set& \bf Accuracy & \bf F1-Score\\
  & \bf (Set \%) & \bf (Set \%)&  & \\
\hline
% Random Sample & - &  ISD(80) & ISD(20) &  5.04& 0.05\\\hline
% Random Forest & - & ISD(80) & ISD(20) & 27.06 & 0.23\\\hline
% SVM & - & ISD(80) & ISD(20) & 21.55 & 0.17\\\hline
% Logistic Reg. & - & ISD(80) & ISD(20) & 27.56 & 0.25\\\hline
ImageNet & PlantDoc (80) & PlantDoc (20)& 13.74 &0.12\\\hline
ImageNet & PVD & PlantDoc (100)&15.08& 0.15\\\hline
ImageNet+PVD & PlantDoc (80) & PlantDoc (20)&\textbf{29.73}&\textbf{0.28}\\ \hline
\end{tabular}}
\end{center}
\caption{Transfer Learning doubled the accuracy after fine-tuning on Uncropped PlantDoc dataset}\label{tab:classes_uncropped}
\vspace{-0.3cm}
\end{table}

\begin{table}[t]
\begin{center}
\resizebox{\hsize}{!}{
\begin{tabular}{c|ccc|cc}
% \hline
\bf Model & \bf PreTrained Weights & \bf Training Set & \bf Test Set& \bf Accuracy & \bf F1-Score\\

 & & \bf (Set \%) & \bf (Set \%) & & \\

\hline
VGG16 & ImageNet & C-PD (80) & C-PD(20)&44.52& 0.44\\\hline
VGG16 & ImageNet & PVD & C-PD (100)&19.73& 0.18 \\\hline
VGG16 & ImageNet+PVD & C-PD (80) & C-PD (20)&60.41& 0.60\\ \hline
InceptionV3 & ImageNet & C-PD (80) & C-PD (20)& 46.67& 0.46\\ \hline
InceptionV3 & ImageNet & PVD & C-PD (100)& 30.78& 0.28 \\ \hline
InceptionV3 & ImageNet+PVD & C-PD (80) & C-PD (20)&62.06& 0.61\\ \hline
InceptionResNet V2 & ImageNet & C-PD (80) & C-PD (20)&49.04& 0.49\\ \hline
InceptionResNet V2 & ImageNet & PVD & C-PD (100)&39.87& 0.38\\ \hline
InceptionResNet V2 & ImageNet+PVD & C-PD (80) & C-PD (20)& \textbf{70.53}& \textbf{0.70}\\ \hline
\end{tabular}}
\end{center}
\caption{Training on controlled dataset (PlantVillage - PVD) gives poor performance on real world images. Performance on real world images can be improved by training on real world images from our dataset}\label{tab:classes_cropped}
\vspace{-0.3cm}
\end{table}

\subsection{System configuration}
All our experiments used NVidia V100 GPU with a system of 32 GB RAM and 8 CPU cores. We used Keras with Tensorflow backend as the deep learning framework. We plan to make the fully reproducible Github repository\footnote{https://github.com/pratikkayal/PlantDoc-Object-Detection-Dataset}$^,$\footnote{https://github.com/pratikkayal/PlantDoc-Dataset} for code and dataset.

\subsection{Plant image classification}
Our main goal was to construct a model which can detect a leaf in an image and then classify it into the particular classes shown in Figure \ref{fig:isd}. We performed two main experiments, which we discuss after describing our experimental settings.
\subsubsection{Experimental settings}
 For training the networks, we used stochastic gradient descent with momentum 0.9, categorical cross-entropy loss, and a learning rate of 0.001. All weights were initialized with the orthogonal initializer. We applied common data augmentation techniques such as rotation, scaling, flipping etc. on the input images. All images were
resized to 100 $\times$ 100, before feeding into the networks. For pre-trained models, we used the weights provided in Keras trained on ImageNet. 

\subsubsection{Plant image classification using raw images (uncropped)}
Our first experiments aims to understand classification accuracy on the uncropped PlantDoc dataset. We evaluated the performance of VGG16 \cite{simonyan2014very} using different training sets on PlantDoc as shown in Table \ref{tab:classes_uncropped}.

\subsubsection{Plant image classification using cropped images}
Further, we evaluate the performance of several popular CNN architectures on the Cropped-PlantDoc dataset that have recently achieved state-of-the-art accuracies on image classification tasks on the popular datasets, such as ImageNet \cite{deng2009imagenet}, CIFAR-10 \cite{krizhevsky2009learning}, etc. Table \ref{tab:classes_cropped} gives the complete list
of the architectures that we used for benchmarking our Cropped-PlantDoc dataset. This experiment was conducted to verify the performance of PlantVillage in real-setting.

\subsection{Leaf Detection}
The aim of our next experiments is to evaluate the performance of Faster R-CNN with InceptionResnetV2 model and MobileNet model on our PlantDoc Dataset as shown in Table \ref{tab:classes_object_detection}. We use mean average precision (mAP: higher is better) to evaluate the models and compare it with scores on COCO dataset since no evaluation exists in the domain of plant disease.

\subsubsection{Experimental Setting}
Object Detection models require training for a much longer duration. 
% For all the models, we took pretrained weights and first loaded it to our model. This drastically reduced the required training time. 
For training Faster R-CNN with Inception Resnet v2 network, we used Momentum optimizer keeping a degrading learning rate with an initial value of 0.0006. For training the MobileNet network, with RMSprop as optimizer -- we took an initial learning rate of 0.0005 with decay steps as 25000 and decay factor as 0.95. While training, data augmentation like random horizontal flip and random SSD crop was applied on input images. We split our dataset into 2,360-238 based on training-testing. We took the pre-trained weights and fine-tuned on training set of PlantDoc. As aforementioned, we provide train-test splits of the dataset for consistent evaluation and fair comparison over the dataset in future.

% Since our main goal was to create a model that could run on mobile devices, we have trained MobileNet models too. These models Employ Single Shot Detection technique, which makes the model computationally efficient.
\begin{table}[t]
\begin{center}
\resizebox{\hsize}{!}{
\begin{tabular}{c|c|c}

\bf Model & \bf PreTrained Weights &\bf mAP (at 50\% iou)\\
\hline
MobileNet & COCO &32.8\\\hline
MobileNet & COCO+PVD &22.4 \\\hline
Faster-rcnn-inception-resnet & iNaturalist &36.1\\ \hline
Faster-rcnn-inception-resnet & COCO & \textbf{38.9}\\ \hline
\end{tabular}}
\end{center} 
\caption{Leaf detection mAP \\$^*$ iou refers to intersection over union}\label{tab:classes_object_detection}
\vspace{-0.3cm}
\end{table}

\begin{figure}[t]
\centering
\begin{tabular}{c}
    \includegraphics[height=2.3cm]{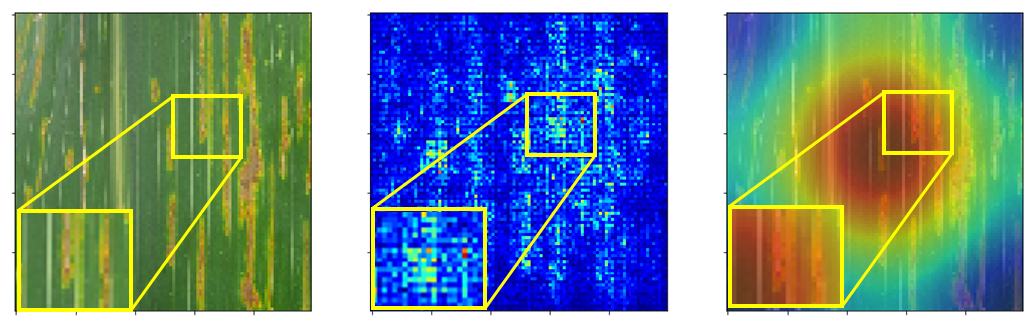}\\
    \includegraphics[height=2.3cm]{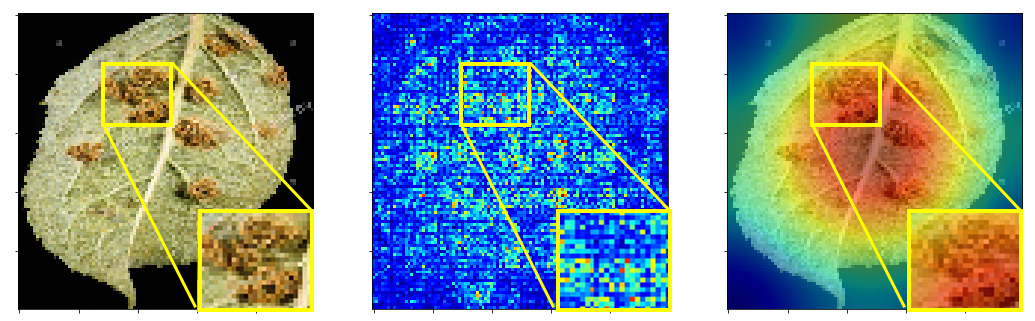}
\end{tabular}
\caption{Saliency and Activation Maps shows the affected parts of disease in a leaf.}\label{fig:gradcam}
\end{figure}

\section{Results and Discussion}
 Figure \ref{fig:gradcam} shows saliency map and gradient activation map of the  Corn Leaf Blight and Tomato Bacterial Spots respectively. As expected, the neural network is learning to focus on the set of visual features which are correlated with disease such as the blemishes in the leaf (lines in Corn Leaf Blight and spots in Tomato Bacterial Spots). The network even learns the shape of the leaf (shown in second row of Figure \ref{fig:gradcam}) to help it distinguish between species.

As predicted, the results in Table \ref{tab:classes_uncropped} clearly shows that real case scenarios have low accuracy when processed initially with ImageNet or PlantVillage. Also, Table \ref{tab:classes_uncropped} and Table \ref{tab:classes_cropped} clearly shows low accuracy achieved by training on PlantVillage and testing on PlantDoc. Model fails to produce accurate results due to background noise, images with leaf from multiple classes in a dataset and low-resolution leaf images. 

% \begin{figure}[!thb]
% \centering
% \resizebox{100}{!}{\begin{tabular}{cc}
%     \includegraphics{fig/Screenshot_20190424-233803.jpg}\\
% \end{tabular}}
% \end{figure}

% \begin{table}[]
% \begin{tabular}{ll}
% \multirow{2}{*}{\includegraphics{fig/Screenshot_20190424-233803.jpg} &  \includegraphics{fig/object.png} \\
%                   & \includegraphics{fig/pasted image 0.png}
% \end{tabular}
% \end{table}

% \begin{figure}[!thb]
% \centering
% \resizebox{\hsize}{!}{\begin{tabular}{cc}
%     \multirow{3}{*}{\includegraphics[height=30cm]{fig/Screenshot_20190424-233803.jpg}}& \includegraphics{fig/object.png}\\
%     &\includegraphics{fig/pasted image 0.png}\\
% \end{tabular}}
% \caption{Object Detection Outputs in Jupyter Notebook}\label{fig:object_1}
% \end{figure}

\begin{figure}
    \centering
    \includegraphics[width=0.27\textwidth]{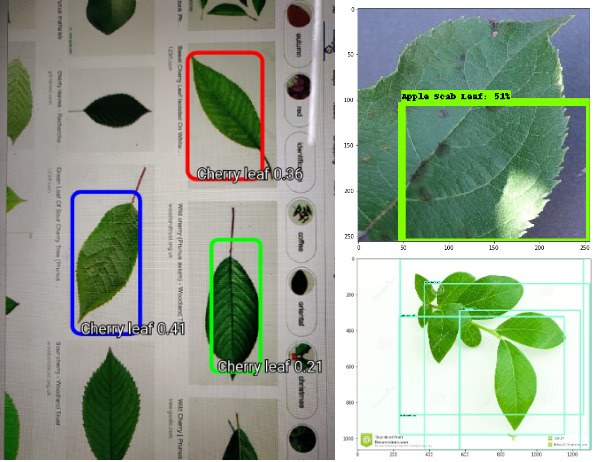}
    \caption{Leaf detection results in our mobile application}
    \label{fig:Object Detection}
\end{figure}

Table \ref{tab:classes_object_detection} shows that Faster R-CNN with InceptionResnetV2 performs the best with an mAP of 38.9. It is interesting to see that MobileNet performance is decreased when pre-trained on COCO+PlantVillage compared to the model where pre-training was done only on COCO. This attributes to the fact that PlantVillage is not contributing towards better results. 
MobileNet gives an mAP of 22 when evaluated on COCO dataset which has significantly more classes \cite{howard2017mobilenets}. 
% Also, these results are comparable to mAP scores of these models on COCO dataset \cite{howard2017mobilenets} although the domain is different and the number of classes are more in COCO.

% Table \ref{tab:classes_cross} shows that the model has reached saturation on Type-3 dataset and shows similar behaviour on the all the splits. This shows that the dataset is well spread and prediction is not able to increase due some common features in different subsets of images. Even Table \ref{tab:classes4} supports this argument as varying test set does not decrease the accuracy drastically. Few examples of the predicted images is shown in Figure \ref{fig:object_1}. 

\begin{figure}[!thb]
\centering
\begin{tabular}{cc}
    \includegraphics[height=2cm]{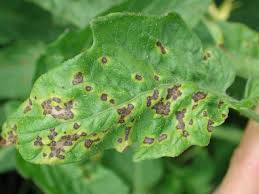}&
    \includegraphics[height=2cm]{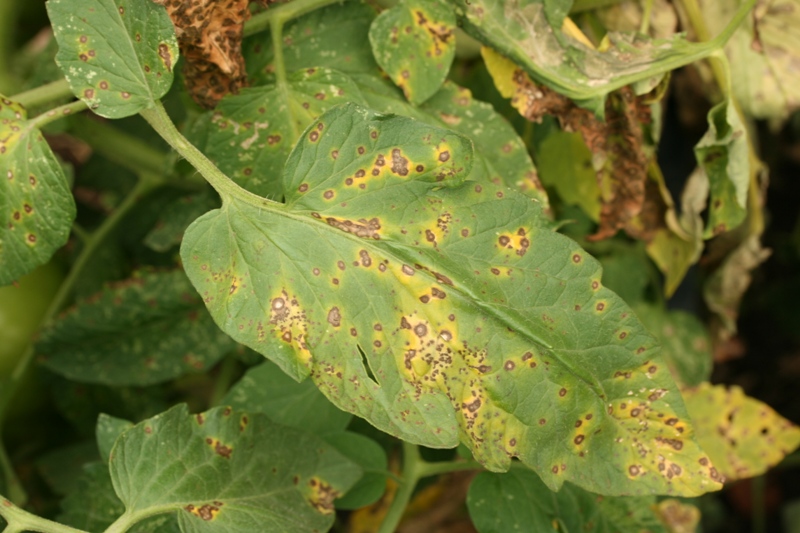}\\
    (a)&(b)
\end{tabular}
\caption{Tomato Bacterial leaf spot(a)and Septoria(b) looks similar and are hard to label using visual features alone}\label{fig:septoria}
\end{figure}

% Figure \ref{fig:confusion} shows the misclassification between the different classes specially the higher misclassification in classes of same species due to similar leaf shape and color and together with Figure \ref{fig:confusion_species}, the misclassification between same disease in different species can be observed. 
% Figure \ref{fig:septoria} shows one such example of high misclassification between septoria(b) and bacterial leaf spot(a) in tomato. Both look very similar to each other and due to lack of domain knowledge, their labelling was very difficult task. 
\section{Application Building}
We were able to adapt the above solution to a mobile environment (Figure \ref{fig:Object Detection}) by using models that very significant reduce complexity, without sacrificing the effective accuracy.
This allowed us to achieve the best possible performance, given that the application should predict the bounding boxes and classes in real time in a mobile CPU. 
We have build application that utilizes MobileNets Object Detection Network ~\cite{DBLP:journals/corr/HowardZCKWWAA17} due to its efficiency and competitive accuracy. The network builds on top of the SSD framework \cite{liu2016ssd}. 
\section{Limitations}
% We now discuss a few limitations of our work. 
The dataset has been curated with care, but due to lack of extensive domain expertise, there are some images in the dataset which can potentially be wrongly classified (shown in Figure \ref{fig:septoria}). Further, to train highly accurate models for disease detection, we may require a dataset with more number of images in each class. But, due to non-availability of public dataset and lack of real-life scenario for field work, our approach gives a feasible direction to tackle the on-going problem of disease detection. 

%\section{Future Work}
% Many different adaptations, tests, and experiments have been left for the future due to lack of time (i.e. the experiments with real data are usually very time consuming, requiring days to finish a single run). 

\section{Conclusions and future work}
In this paper, we addressed the problem of detection of diseased/healthy leaves in images using state of the art object detection models. One of the main contributions of our work is to propose
an entirely new dataset for plant disease detection called PlantDoc. Our benchmark experiments show the lack of efficacy of models learnt on controlled datasets, thereby, showing the significance of real-world datasets such as ours. Applying image segmentation techniques to extract leaf out of the images can potentially enhance the utility of the dataset. We believe that this dataset is an important first step towards computer vision enabled scalable plant disease detection. 
\clearpage

\bibliographystyle{ACM-Reference-Format}
\bibliography{ref}

\end{document}